\documentclass{article}

\usepackage[preprint]{neurips_2025}

\usepackage[utf8]{inputenc} 
\usepackage[T1]{fontenc}    
\usepackage{hyperref}       
\usepackage{url}            
\usepackage{booktabs}       
\usepackage{amsfonts}       
\usepackage{nicefrac}       
\usepackage{microtype}      
\usepackage{xcolor}         

\usepackage{amsmath}
\usepackage{array}  
\usepackage{bm}
\usepackage{graphicx}
\usepackage{caption} 
\usepackage{multirow}

\definecolor{darkblue}{RGB}{0,0,139}
\definecolor{darkred}{RGB}{139,0,0}

\usepackage{makecell}
\usepackage{pifont}
\newcommand{\cmark}{\textcolor{green!60!black}{\ding{51}}} 
\newcommand{\xmark}{\textcolor{red}{\ding{55}}}            

\newcolumntype{C}{>{\centering\arraybackslash}m{1.5cm}}
\newcommand{\imgicon}{\includegraphics[height=1em]{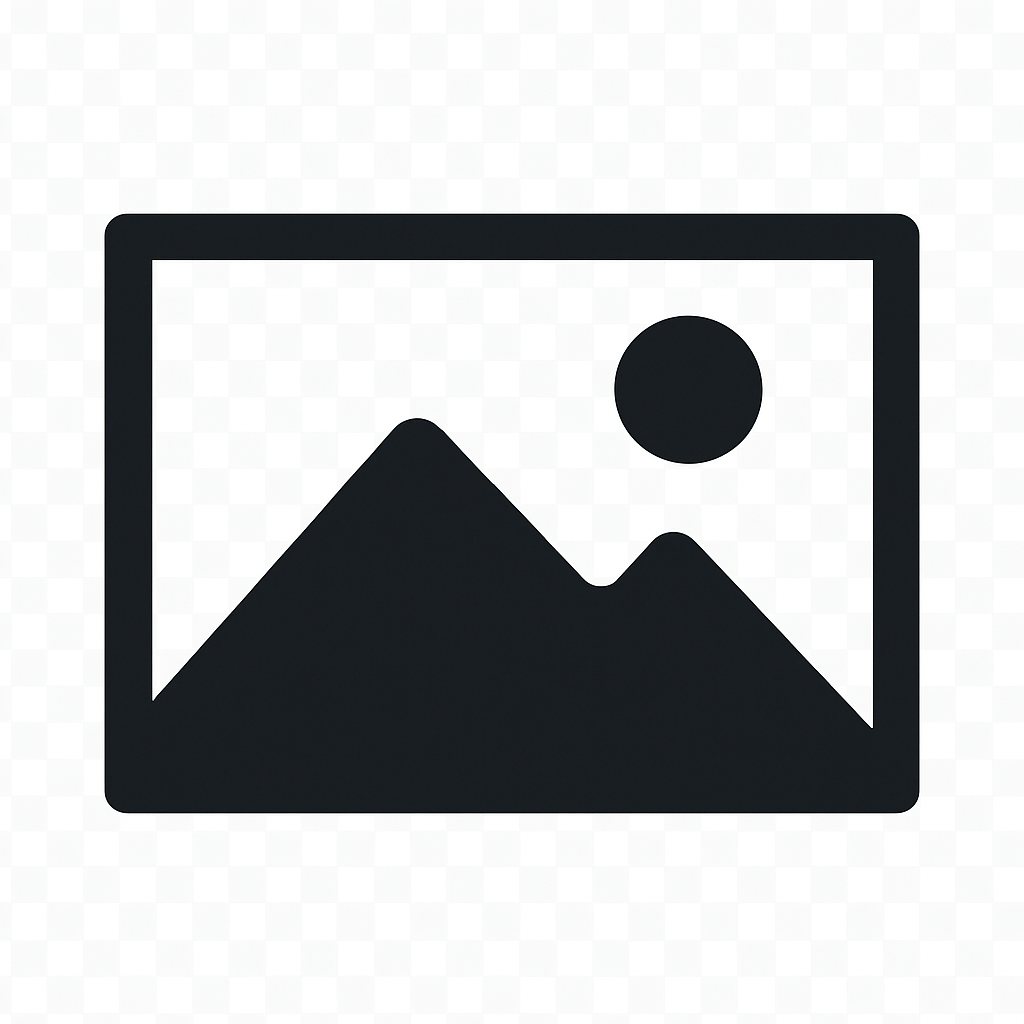}}
\newcommand{\vidicon}{\includegraphics[height=1em]{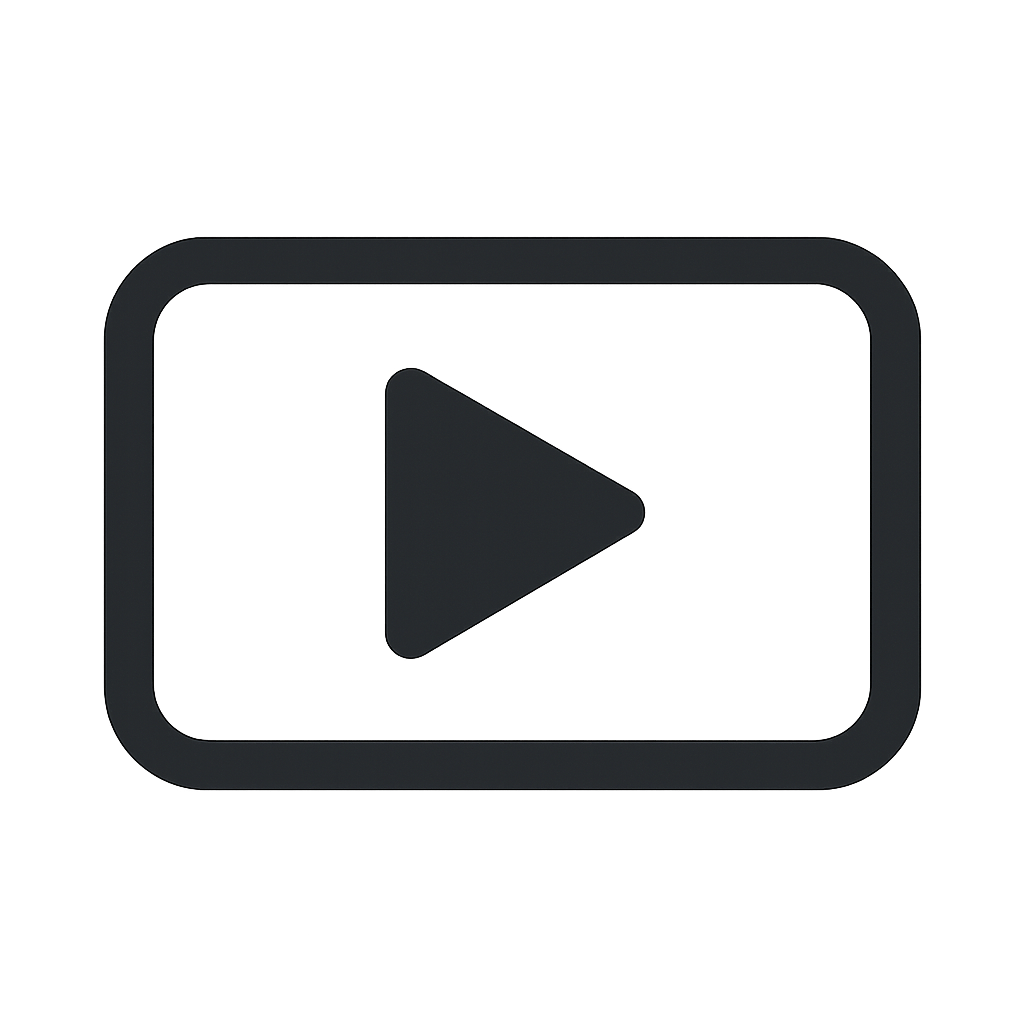}}

\newcommand{\plh}{---}             

\title{RVTBench: A Benchmark for Visual Reasoning Tasks
}

\author{%
Yiqing Shen, Chenjia Li, Chenxiao Fan, Mathias Unberath\\
Johns Hopkins University
}

\begin{document}

\maketitle

\begin{abstract}

Visual reasoning, the capability to interpret visual input in response to implicit text query through multi-step reasoning, remains a challenge for deep learning models due to the lack of relevant benchmarks.
Previous work in visual reasoning has primarily focused on reasoning segmentation, where models aim to segment objects based on implicit text queries.
This paper introduces reasoning visual tasks (RVTs), a unified formulation that extends beyond traditional video reasoning segmentation to a diverse family of visual language reasoning problems, which can therefore accommodate multiple output formats including bounding boxes, natural language descriptions, and question-answer pairs.
Correspondingly, we identify the limitations in current benchmark construction methods that rely solely on large language models (LLMs), which inadequately capture complex spatial-temporal relationships and multi-step reasoning chains in video due to their reliance on token representation, resulting in benchmarks with artificially limited reasoning complexity.
To address this limitation, we propose a novel automated RVT benchmark construction pipeline that leverages digital twin (DT) representations as structured intermediaries between perception and the generation of implicit text queries. 
Based on this method, we construct RVTBench, a RVT benchmark containing 3,896 queries of over 1.2 million tokens across four types of RVT (segmentation, grounding, VQA and summary), three reasoning categories (semantic, spatial, and temporal), and four increasing difficulty levels, derived from 200 video sequences. 
Finally, we propose RVTagent, an agent framework for RVT that allows for zero-shot generalization across various types of RVT without task-specific fine-tuning.
Dataset and code are available at \url{https://huggingface.co/datasets/yiqingshen/rvtbench/tree/main/rvtbench} and \url{https://github.com/yiqings/rvt}.
\end{abstract}

\section{Introduction}
Visual understanding combined with reasoning is important for various applications, such as in embodied AI or human-computer interaction, in the interpretation of complex real-world scenarios.
Although previous progress has been made in visual perception through various visual foundation models such as SAM \cite{sam1,sam2} and DINO \cite{dino}, these models primarily excel at recognizing what is present in a scene rather than reasoning about it.
For example, while current models can identify cups, tables, and people in an image with high accuracy, they struggle with requests like ``\textit{bring me something to pour coffee into}'' or ``\textit{find the object that the person on the left will interact next}'', which are tasks that require both perception and reasoning.
Reasoning segmentation makes the first exploration of this direction by segmenting objects from images or videos based on \textbf{implicit text queries} \cite{lisa,visa}, which refer to queries that do not directly describe the target object but instead require inference about its properties, functions, or relationships to identify it.
Unlike traditional task formulations such as semantic segmentation with predefined categories or referring segmentation \cite{referringsegmentation} with explicit object descriptions, reasoning segmentation requires models to process both the visual data and complex text queries through multi-step reasoning to identify target objects. 
For example, instead of responding to queries such as ``\textit{segment the coffee cup},'' reasoning segmentation handles implicit text queries such as ``\textit{segment the object used for holding hot beverages},'' that require both visual perception and semantic reasoning about object functionality.

\begin{figure*}[htbp!]
\centering
\includegraphics[width=0.8\linewidth]{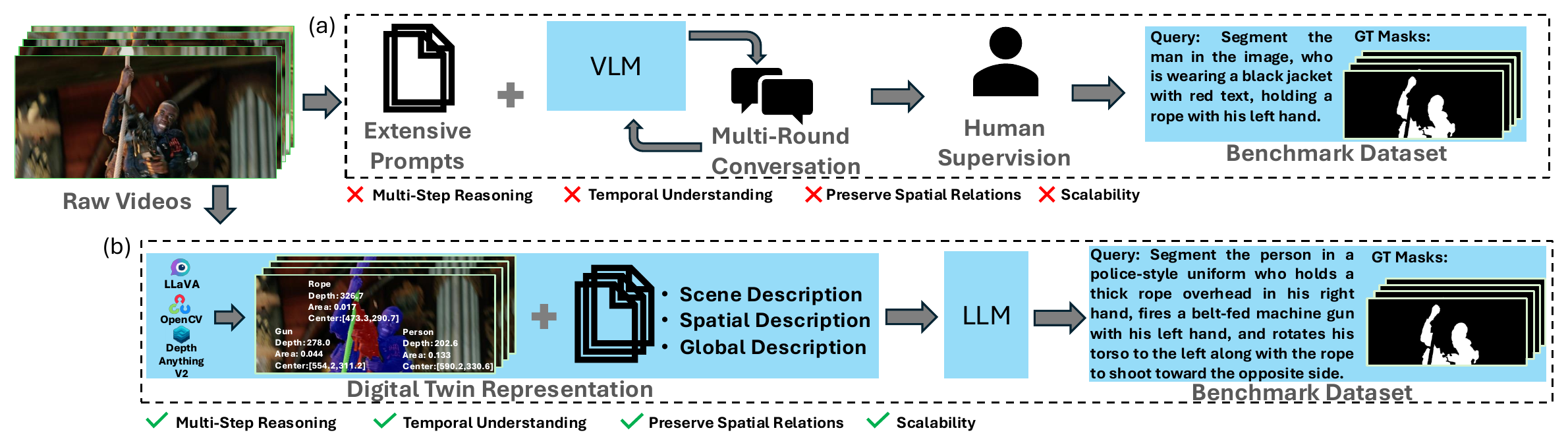}
\caption{
Comparison of benchmark dataset construction approaches for reasoning visual tasks. 
(a) Traditional VLM-based approach using extensive prompts and human supervision, which struggles with multi-step reasoning, temporal understanding, spatial relations preservation, and scalability. 
(b) Our proposed digital twin representation approach that leverages specialized vision foundation models to create structured intermediate representations enabling more complex implicit text queries.
}\label{fig:intro}
\end{figure*}

However, reasoning segmentation is constrained by its singular focus on the pixel-level segmentation mask as the output format.
In practice, real-world visual tasks can require diverse response formats depending on the context and application needs. 
For instance, object detection may need bounding boxes for efficient localization, human-AI collaborative frameworks often benefit from natural language descriptions of visual content, and question-answer workflows may generate text responses about visual data \cite{shen2025operating}. 
This limitation of the output mode of reasoning segmentation restricts the broader applicability in scenarios that require more versatile visual reasoning capabilities.
To address these constraints, we propose \textbf{reasoning visual tasks} (RVTs), a unified formulation that generalizes reasoning segmentation into a family of vision language problems. 
RVT preserves the fundamental principle of processing implicit reasoning-driven text queries while expanding the output format to encompass more diverse tasks, including reasoning segmentation to produce pixel-level masks \cite{lisa}, reasoning grounding to generate bounding boxes \cite{zhu2024scanreason}, reasoning summary to provide descriptions in natural language, and reasoning visual question answer to offer textual answers \cite{shen2025operating}. 
This formulation not only offers greater flexibility in how models respond to implicit queries, but also enables more intuitive human-machine interactions across domains such as autonomous navigation, medical image analysis \cite{shen2025operating}, and augmented reality applications.

Developing standardized benchmark datasets is important for advancing RVTs, as they enable objective performance evaluation and facilitate fair comparison between methods.
Despite this importance, existing benchmarks for RVT often suffer from two limitations, namely (1) a narrow focus on single output modalities (predominantly segmentation) and (2) insufficient complexity in the reasoning chains required to solve them.
Moreover, current automated benchmark dataset construction approaches for RVT rely on large language models (LLMs) or vision language models (VLMs) to generate implicit queries \cite{lisa,llmseg}, as shown in Fig.~\ref{fig:intro}.
For example, in terms of image reasoning segmentation, LLM-Seg40K \cite{llmseg} uses a two-stage pipeline to generate image reasoning segmentation datasets, where LLaVA \cite{llava} first generates detailed image descriptions that GPT-4 later transforms into implicit queries.
Similarly, ReasonSeg-Ins \cite{lisa++} uses GPT-4V to directly generate implicit text queries and answer pairs from images. 
For video reasoning segmentation, VideoReasonSeg \cite{villa} uses GPT-4V to analyze videos with instance annotations to generate question-answer pairs requiring temporal reasoning, while ReasonVOS \cite{videolisa} employs LLMs to rephrase and augment explicit text queries from referring segmentation datasets.
However, these approaches face three major limitations due to their reliance on token-based representations in LLMs and VLMs, where the continuous nature of visual-spatial-temporal relationships is fragmented into discrete tokens \cite{shen2025position}.
First, VLMs struggle to encode complex spatial relationships effectively, as tokenization discretizes spatial continuity into fixed-length tokens that lose fine-grained positional information, resulting in generated queries with oversimplified spatial reasoning and inconsistent geometric understanding \cite{shen2025position}. 
Second, VLMs inadequately represent the temporal dynamics between video frames by compressing sequential information into simplified token sequences, producing queries that lack complex temporal reasoning \cite{shen2025position}.
Finally, as a consequence of these spatial and temporal limitations, VLMs struggle to generate queries requiring multi-step reasoning, resulting in benchmark datasets with artificially limited reasoning complexity that fail to evaluate models on the deep inferential capabilities needed for real-world applications \cite{shen2025position}.
To bridge the gap, we propose a novel benchmark dataset construction approach leveraging digital twin (DT) representations, defined as ``\textit{a paradigm that creates outcome-driven digital replicas of physical processes that capture task-specific entities and their interactions}'' \cite{fuller2020digital,shen2025position}. 
Specifically, DT representations can serve as intermediaries between the perception of raw visual data and high-level reasoning by maintaining explicit entity relationships that preserve the continuous nature of visual information \cite{shen2025position}. 
Unlike token-based representations that fragment spatial-temporal relationships, our DT representation approach explicitly models semantic categories, spatial geometries, and temporal dynamics \cite{shen2025position}. 
Table \ref{table:datasets_full} presents a comparison of our proposed dataset with existing ones.

The major contributions are four-fold.
First, we formally define reasoning visual tasks as a unified family of visual language problems that require both visual perception and reasoning over implicit text queries. 
This formulation generalizes reasoning segmentation to accommodate multiple output formats, including segmentation masks, bounding boxes, natural language summaries, and question-answer pairs.
Second, we propose an automated benchmark dataset construction pipeline that leverages DT representations that decouple perception from reasoning. 
Unlike previous approaches that rely solely on VLMs or LLMs to generate implicit queries, our method enables more precise control over implicit text query complexity while ensuring alignment with ground-truth annotations without human intervention.
Third, based on the previous automated benchmark dataset construction method, we introduce RVTBench, an RVT benchmark dataset comprising 3,896 queries in four types of RVT, three reasoning categories, and four difficulty levels from 200 videos. 
Fourth, we present RVTagent, a baseline method for RVT that does not require task-specific fine-tuning. 

\begin{table*}[t]
\centering
\caption{
Comprehensive comparison of reasoning visual task benchmarks across multiple dimensions. 
For each dataset, we compare input modalities (image \imgicon or video \vidicon), implicit text query characteristics (multi-level complexity, reasoning categories, generation method), annotation approaches (mask source and whether ground truth is automatically created), supported reasoning task types (segmentation, grounding, VQA, summary), and dataset scale metrics (image/video count and queries). 
Unlike previous benchmarks that focus primarily on segmentation with limited reasoning complexity, RVTBench uniquely supports all four task types while providing comprehensive coverage of semantic, spatial, and temporal reasoning at multiple difficulty levels.
}
\label{table:datasets_full}
\resizebox{\textwidth}{!}{%
\begin{tabular}{l|C|ccccc|cc|cccc|ccc}
\hline
\multirow{2}{*}{\textbf{Benchmark Dataset}}
& \multirow{2}{*}{\textbf{Modalities}}
& \multicolumn{5}{c|}{\textbf{Type of Implicit Text Query}}
& \multicolumn{2}{c|}{\textbf{Annotation Approach}}
& \multicolumn{4}{c|}{\textbf{Type of Reasoning Visual Task}}
& \multicolumn{3}{c}{\textbf{Dataset Scale}} \\
\cline{3-7} \cline{8-9} \cline{10-13} \cline{14-16}
&
& \textbf{Multi-lvl} & \textbf{Semantic} & \textbf{Spatial}
& \textbf{Temporal} & \textbf{Generation}
& \textbf{Mask Annotation} & \textbf{Auto GT}
& \textbf{Seg} & \textbf{Grounding} & \textbf{VQA} & \textbf{Summary}
& \textbf{Image} & \textbf{Video} & \textbf{Query} \\
\hline
LLM-Seg40K \cite{llmseg}
& \imgicon
& \xmark & \cmark & \xmark & \xmark & LLM Generated
& SAM \cite{sam1} + Src & \cmark
& \cmark & \xmark & \xmark & \xmark
& 14,000 & --     & 55,300 \\
ReasonSeg-Ins \cite{lisa++}
& \imgicon
& \xmark & \cmark & \cmark & \xmark & LLM Generated
& Src & \cmark
& \cmark & \xmark & \xmark & \xmark
& 63,800 & --     & 63,800 \\
VideoReasonSeg \cite{villa}
& \vidicon
& \cmark & \xmark & \cmark & \xmark & LLM Modified
& Src & \xmark
& \cmark & \xmark & \xmark & \xmark
& --    & 1,934  & 21,000 \\
ReasonVOS \cite{videolisa}
& \vidicon
& \xmark & \xmark & \xmark & \xmark & LLM Modified
& Src & \xmark
& \cmark & \xmark & \xmark & \xmark
& --    & 91     & 458 \\
ReVOS \cite{visa}
& \vidicon
& \xmark & \cmark & \cmark & \xmark & Human Annotated
& Src & \xmark
& \cmark & \xmark & \xmark & \xmark
& --    & 1,042  & 35,074 \\
JiT Bench \cite{jit}
& \vidicon
& \cmark & \cmark & \cmark & \cmark & Human Annotated
& Src & \xmark
& \cmark & \xmark & \xmark & \xmark
& --    & 200    & 895 \\
GroundMORE \cite{mora}
& \vidicon
& \cmark & \cmark & \cmark & \xmark & LLM Modified
& XMem++ \cite{bekuzarov2023xmemproductionlevelvideosegmentation} & \cmark
& \cmark & \xmark & \xmark & \xmark
& --    & 1,715  & 7,577 \\
\hline
RVTBench (Ours)
& \vidicon
& \cmark & \cmark & \cmark & \cmark & LLM Generated
& SAM2 \cite{sam2} + Src & \cmark
& \cmark & \cmark & \cmark & \cmark
& 63,463 & 200    & 3,896 \\
\hline
\end{tabular}%
}
\end{table*}

\section{Methods}
\subsection{Problem Definition}
We define \textbf{reasoning visual tasks} (RVT) as a family of vision-language problems that require perception of visual data and reasoning over implicit text queries. 
This can include reasoning segmentation (producing pixel-level masks) \cite{lisa}, reasoning grounding (generating bounding boxes) \cite{zhu2024scanreason}, reasoning summary (generating textual descriptions of objects), reasoning visual question answering (VQA, providing natural language answers), and others.
In this formulation, the RVT model must determine not only \textit{what} visual elements to focus on, but also \textit{how} to process them based on an implicit query.
Formally, given an input video $\mathcal{X} = \{I^{(1)}, I^{(2)}, \ldots, I^{(T)}\} \in \mathbb{R}^{T\times H \times W \times 3}$ consisting of $T$ frames and a text query $Q$ that implicitly describes the goal, RVT aims to produce the corresponding output $\mathcal{Y}$ through a reasoning process $\mathcal{R}$. 
We focus on video because it naturally generalizes to static images, which can be treated as single-frame videos, while enabling the evaluation of temporal reasoning capabilities. 
Typically, the reasoning process $\mathcal{R}$ can be decomposed into two stages: 
\begin{equation} 
\mathcal{Y} = \mathcal{R}(\mathcal{X}, Q) = \mathcal{E}(\mathcal{T}(Q), \mathcal{X}, Q),
\end{equation}
where $\mathcal{T}$ is the task identification function that determines the appropriate visual operation to perform based on the query $Q$, and $\mathcal{E}$ represents the task execution function that applies the identified operation to relevant objects in $\mathcal{X}$ as implicitly specified by $Q$.
In terms of reasoning segmentation task, $\mathcal{Y} = \{M^{(1)}, M^{(2)}, \ldots, M^{(T)}\}$ represents a sequence of binary segmentation masks, where each $M^{(t)} \in \{0, 1\}^{H \times W}$ indicates the pixels that satisfy $Q$ in frame $I^{(t)}$. 
For reasoning grounding task, $\mathcal{Y} = \{B^{(1)}, B^{(2)}, \ldots, B^{(T)}\}$ becomes a sequence of bounding boxes, where each $B^{(t)} = \{(x_i, y_i, w_i, h_i)\}_{i=1}^{N_t}$ localizes $N_t$ objects in frame $I^{(t)}$ that fulfill $Q$. 
For the reasoning summary task, $\mathcal{Y} = S$ is a natural language summary that describes visual content that captures relevant visual elements and their relationships implicitly specified in $Q$ throughout the temporal dimension. 
For the reasoning VQA task, $\mathcal{Y} = A$ is a natural language answer to the reasoning query $Q$ based on visual information in $\mathcal{X}$. 
What distinguishes the RVTs from their traditional counterparts \cite{referringsegmentation} is the complexity of the reasoning process $\mathcal{R}$. 
Traditional ones typically rely on explicit instructions (\textit{e}.\textit{g}., ``\textit{segment the dog}'') or predefined categories, whereas RVTs handle queries requiring multi-step inference (\textit{e}.\textit{g}., ``\textit{identify the animal that initially appears from the left side of the frame and later interacts with the person wearing red}'').

Following previous work \cite{jit}, the reasoning process $\mathcal{R}$ can be further decomposed into three categories according to the nature of the reasoning required. 
First, semantic reasoning involves understanding the attributes, categories, and relationships of objects based on world knowledge.
Then, spatial reasoning focuses on understanding the relative positions and geometric relationships between objects.
Finally, temporal reasoning refers to understanding motion, sequences, and events over time.
Often, complex reasoning queries of RVTs can involve multiple categories of reasoning.

\begin{figure*}[t!]
\centering
\includegraphics[width=\linewidth]{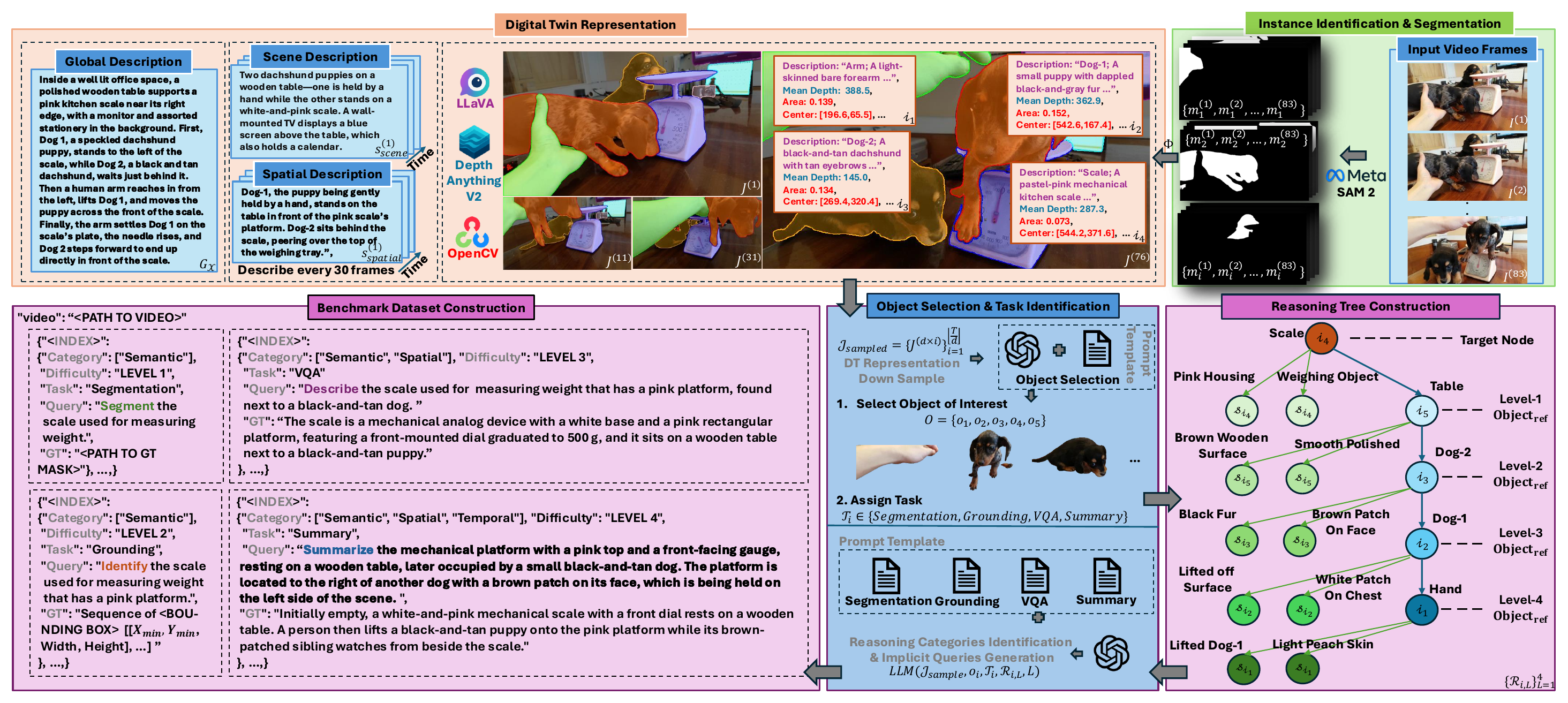}
\caption{
Overview of our automated benchmark dataset construction pipeline for reasoning visual tasks. 
The pipeline includes three components to generate complex data samples without human intervention: (1) Digital twin representation construction, where specialized vision foundation models extract multi-dimensional information from input video frames—including global descriptions, scene-level semantic context, spatial relationships, and instance-specific attributes with depth statistics. This creates a structured JSON that preserves continuous visual-spatial-temporal relationships. 
(2) Object selection and reasoning tree construction, which first identifies objects of interest from down-sampled DT representations, assigns appropriate task types, and then builds a hierarchical reasoning graph with increasing complexity levels (L1-L4). 
Each level progressively incorporates more complex relationships between target objects and their attributes. 
(3) Benchmark dataset construction, which leverages the reasoning tree to generate task-specific implicit queries with corresponding ground-truth annotations at varying difficulty levels, incorporating semantic, spatial, and temporal reasoning categories. 
    }\label{fig:framework}
\end{figure*}

\subsection{Benchmark Dataset Construction}

\paragraph{Data Structure}
Each sample in our benchmark is represented as a tuple 
$\mathcal{D} = \{\mathcal{X}, \mathcal{T}, \mathcal{Y}, \mathcal{J}, Q, C, L\}$, 
where $\mathcal{X}$ is the source video sequence, $\mathcal{T}$ indicates the specific type of RVT with $\mathcal{T} \in \{\text{segmentation}, \text{grounding}, \text{summary}, \text{VQA}\}$, $\mathcal{Y}$ is the corresponding ground truth output with respect to $\mathcal{T}$, $\mathcal{J}$ denotes the corresponding DT representation for $\mathcal{X}$, $Q$ is the implicit text query, $C \subseteq \{\text{semantic}, \text{spatial}, \text{temporal}\}$ specifies the reasoning categories of $Q$, and $L$ defines the difficulty level from 1 to 4 based on the complexity of the reasoning chain. 

\paragraph{Video Data Source}
Our benchmark leverages video sequences from two complementary datasets.
We utilized 62 videos from DAVIS \cite{davis}, which consists of carefully captured full HD sequences that feature multiple instances of common video object segmentation challenges such as occlusions, motion blur, and appearance changes.
Furthermore, we incorporate 138 videos from SA-V \cite{sam2}, collected by crowdworkers in 47 countries that capture indoor and outdoor scenes.
Note that we only adopt the raw video sequences from these datasets, without using their provided segmentation masks.

\paragraph{Method Overview}
We propose an automated benchmark data set construction pipeline that uses digital twin representations and LLM to generate RVT datasets.
Unlike previous benchmark dataset generation approaches in reasoning segmentation that rely solely on VLMs to generate implicit queries or LLMs to rephrase referring queries \cite{villa,videolisa,athar2024vicas}, which often struggle with coming up with a complex query involving spatial relationships and temporal reasoning \cite{jit}; our approach decouples perception from reasoning with DT representation, enabling more precise control over implicit text query complexity and providing the corresponding ground-truth annotations without human intervention.
Our pipeline consists of three stages.
First, we transform the input video sequence $\mathcal{X}$ into a structured DT representation $\mathcal{J}$ to preserve the semantic, spatial, and temporal relationships of objects with specialized vision foundation models including SAM2 \cite{sam2} for instance identification and segmentation, DepthAnythingv2 \cite{depthanything} for depth estimation, VLM \cite{llava} for instance-level and scene-level descriptions, and conventional OpenCV-based operators for frame-level processing.
This DT representation encodes objects with their attributes, positions, and temporal relationships in a JSON structure.
Second, objects of interest are randomly chosen by sampling from the DT representation, and then we prompt the LLM to identify the appropriate specific task type $\mathcal{T}$ with respect to this object, which will be later used to generate the corresponding implicit text query. 
Third, we construct a reasoning tree from this DT representation $\mathcal{J}$ that hierarchically organizes the object information at different levels of abstraction and progressively generates implicit text queries $Q$ of different levels of difficulty with respect to $\mathcal{T}$.
The reasoning tree is structured as a directed acyclic graph (DAG) with nodes representing objects and edges indicating relationships.
Finally, each query is $Q$ paired with the corresponding ground truth retrieved from the DT representation.

\paragraph{Digital Twin Representation Construction}
For each video sequence $\mathcal{X} = \{I^{(1)}, I^{(2)}, \ldots, I^{(T)}\}$, we construct a corresponding DT representation $\mathcal{J} = \{J^{(1)}, J^{(2)}, \ldots, J^{(T)}\}$, where each $J^{(t)}$ encodes frame-level information in the timestep $t$. 
This DT representation serves as a structured intermediate layer that bridges raw visual data and high-level reasoning processes in subsequent data generation.
The DT construction process employs a suite of specialized vision foundation models $\Phi = \{\phi_1, \phi_2, ..., \phi_K\}$ to extract information, formally expressed as $J^{(t)} = \Phi(I^{(t)})$. 
To balance computational efficiency with temporal coherence, we process key-frames with $\Phi$ at intervals of $t_s$ frames and propagate information to intermediate frames.
Firstly, we utilize SAM2 to generate instance segmentation masks $M^{(t)} = \{m_i^{(t)}\}_{i=1}^{N^{(t)}}$, where each $m_i^{(t)}$ represents a binary mask for object $i$ with confidence score $\beta_i^{(t)}$. 
For frames between key frames, we leverage SAM2's memory-based tracking to maintain consistent instance identification:
\begin{equation}
m_i^{(t+k)} = \text{SAM}_{\text{track}}(I^{(t+k)}, \{ m_i^{(t+k')}\}_{k'=0}^k), \quad 0 < k < t_s
\end{equation}
To enable better spatial reasoning, DepthAnythingv2 generates depth maps $D^{(t)}$ for each frame. 
For every instance $i$, we compute the depth statistics $d_i^{(t)} = \{D^{(t)}(p) \mid p \in m_i^{(t)}\}$ across all pixels $p$ within its instance mask $m_i^{(t)}$. 
These continuous depth values are summarized in LLM-processable statistics, including mean depth $\mu_i^{(t)} = \frac{1}{|m_i^{(t)}|}\sum_{p \in m_i^{(t)}} D^{(t)}(p)$ and standard deviation $\sigma_i^{(t)} = \sqrt{\frac{1}{|m_i^{(t)}|}\sum_{p \in m_i^{(t)}} (D^{(t)}(p) - \mu_i^{(t)})^2}$.
For semantic understanding, LLaVA-v1.6 \cite{llava} generates both instance-level descriptors $S^{(t)} = \{s_i^{(t)}\}_{i=1}^{N^{(t)}}$ that capture object attributes and categories, and frame-level scene descriptions $s_{\text{scene}}^{(t)}$ that summarize the environment, weather conditions, crowd activity, and identifiable location features in a concise and coherent paragraph. 
We additionally generate spatial descriptions $s_{\text{spatial}}^{(t)}$ that encode the relative positioning of objects within each frame using natural language descriptors with VLM such as ``\textit{front},'' ``\textit{back},'' and ``\textit{next to}'' 
These spatial relationships are derived from the depth statistics ($\mu_i^{(t)}$, $\sigma_i^{(t)}$) and the center coordinates of each instance, where objects with similar depth values (difference $\leq$ 10) are considered approximately the same distance from the viewpoint. 
Spatial descriptions avoid numerical values and instead focus on qualitative relationships between objects, enabling more effective reasoning about relative positions.
In addition to these, conventional OpenCV operators extract additional visual features $V^{(t)} = \{v_i^{(t)}\}_{i=1}^{N^{(t)}}$ including color histograms, optical flow vectors for motion tracking, and texture descriptors.
Finally, to capture video-level context, we generate a global description $G_{\mathcal{X}}$ by applying LLaVA-Video to sampled key frames $G_{\mathcal{X}} = \text{VLM}(\{I^{(k \cdot t_s)}\}_{k=0}^{\lfloor T/t_s \rfloor})$.
The complete DT representation is organized in a JSON structure with three levels: (1) video-level metadata, (2) frame-level information, and (3) instance-level attributes, formally:
\begin{equation}
\mathcal{J} = \left\{
\begin{array}{l}
\textcolor{darkblue}{\text{"metadata"}}: \{\textcolor{darkblue}{\text{"description"}}: \textcolor{darkred}{G_{\mathcal{X}}}, \textcolor{darkblue}{\text{"duration"}}: \textcolor{darkred}{T}, \textcolor{darkblue}{\text{"resolution"}}: \textcolor{darkred}{[H, W]}\}, \\
\textcolor{darkblue}{\text{"frames"}}: \{\textcolor{darkred}{J^{(1)}}, \textcolor{darkred}{J^{(2)}}, \ldots, \textcolor{darkred}{J^{(T)}}\}
\end{array}
\right\}
\end{equation}
where each frame entry $J^{(t)}$ contains:
\begin{equation}
J^{(t)} = \left\{
\begin{array}{l}
\textcolor{darkblue}{\text{"timestamp"}}: \textcolor{darkred}{t}, \; 
\textcolor{darkblue}{\text{"scene\_description"}}: \textcolor{darkred}{s_{\text{scene}}^{(t)}},\; 
\textcolor{darkblue}{\text{"spatial\_description"}}: \textcolor{darkred}{s_{\text{spatial}}^{(t)}},\\
\textcolor{darkblue}{\text{"instances"}}: \{\textcolor{darkred}{i_1}: \{\textcolor{darkblue}{\text{"mask"}}: \textcolor{darkred}{m_{i_1}^{(t)}}, \textcolor{darkblue}{\text{"depth\_stats"}}: [\textcolor{darkred}{\mu_{i_1}^{(t)}}, \textcolor{darkred}{\sigma_{i_1}^{(t)}}], \\
\quad \quad \quad \quad \quad \quad \textcolor{darkblue}{\text{"description"}}: \textcolor{darkred}{s_{i_1}^{(t)}}, \textcolor{darkblue}{\text{"visual\_features"}}: \textcolor{darkred}{v_{i_1}^{(t)}}\}, \ldots\}
\end{array}
\right\}.
\end{equation}

\begin{figure*}[t!]
\centering
\includegraphics[width=\linewidth]{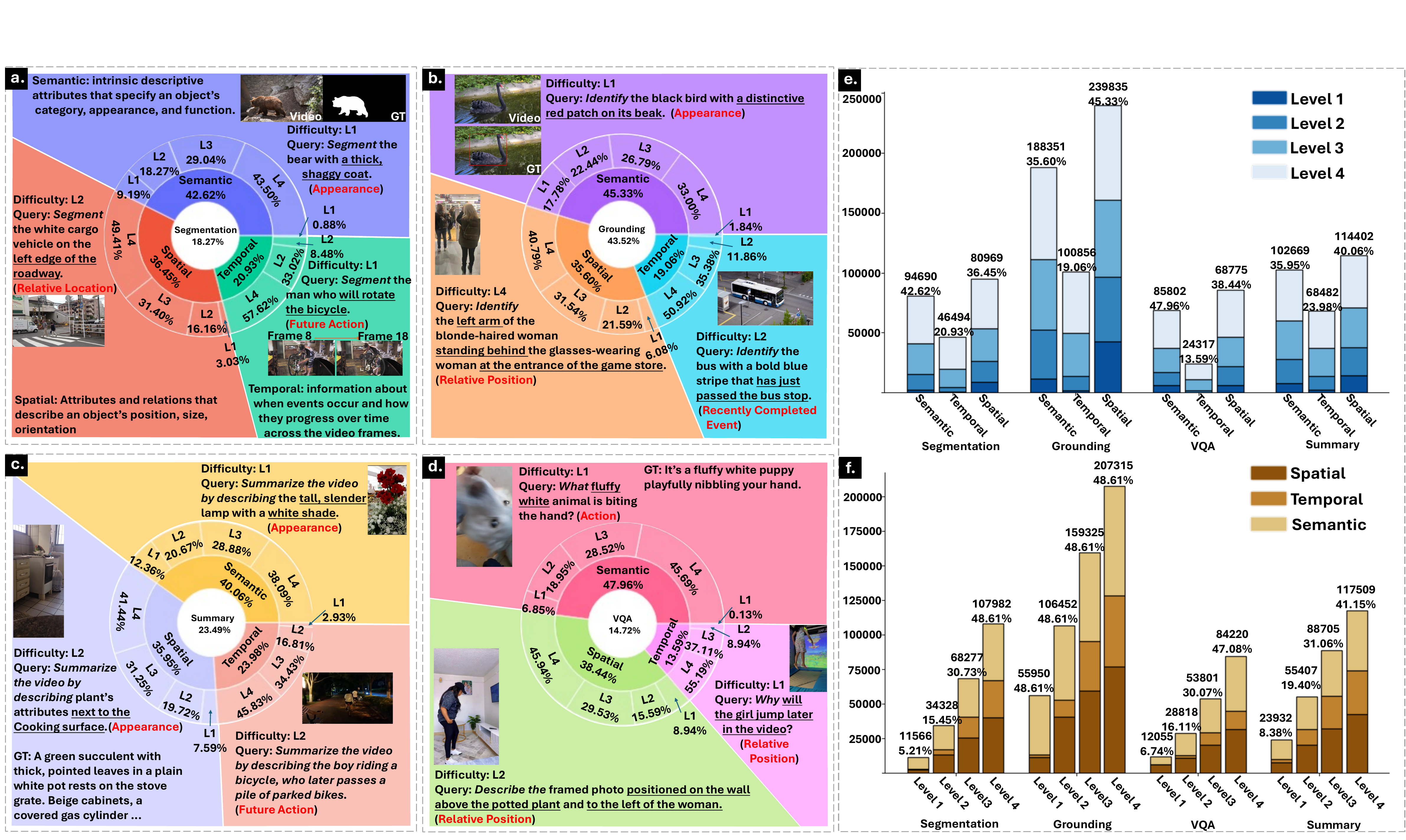}
\caption{
Visualization of the RVTBench composition and examples across different dimensions. 
(a-d) Task-specific sunburst charts illustrating the distribution of queries across reasoning categories (semantic, spatial, temporal) and difficulty levels (L1-L4) for segmentation, grounding, summary, and VQA tasks. 
Each chart includes representative examples that demonstrate the progression in reasoning complexity, from simple attribute identification at L1 (\textit{e}.\textit{g}., ``\textit{Segment the bear with a thick, shaggy coat}'') to complex multi-step reasoning chains at L4 (\textit{e}.\textit{g}., ``\textit{Identify the left arm of the blonde-haired woman standing behind the glasses-wearing woman at the entrance of the game store}'').
Reasoning categories are color-coded (semantic: blue, spatial: red, temporal: green) with annotations highlighting specific reasoning types. 
(e) Token distribution analysis by task type and reasoning category, revealing. (f) Hierarchical breakdown of token distribution across difficulty levels and reasoning categories for each task type. 
}\label{fig:stats}
\end{figure*}

\paragraph{Dataset Generation with LLM}
The overall workflow is shown in Fig.~\ref{fig:framework}.
Given the DT representation $\mathcal{J}$, we employ LLMs to automatically generate RVT queries and the corresponding ground truth. 
To reduce computational overhead, we first down-sample the DT representation by selecting key frames with fixed intervals of $d$, that is, $\mathcal{J}_{\text{sampled}} = \{J^{(d\times i)}\}_{i=1}^{\lfloor T/d \rfloor}$. 
Using this condensed representation $\mathcal{J}_{\text{sampled}}$, we prompt the LLM to identify objects of interest to obtain a ranked list of candidate objects $O = \{o_1, o_2, \ldots, o_N\}$. 
For each selected object $o_i$, we determine the most appropriate task type $\mathcal{T}_i$ through LLM about the characteristics of the object, namely segmentation tasks for objects with distinctive boundaries and meaningful parts, grounding tasks for objects with important spatial positioning, summary tasks for objects with rich semantic attributes or narrative relevance, and VQA tasks for objects involved in complex interactions or state changes. 
We then construct a comprehensive reasoning tree $\mathcal{R}_i = (V_i, E_i)$ for each object $o_i$, where the nodes $V_i$ represent entities, and the edges $E_i$ capture the relationships between them. 
From this reasoning tree, we derive four sub-trees $\{\mathcal{R}_{i,L}\}_{L=1}^4$ sharing the same root node $o_i$ but with increasing depths corresponding to complexity levels $L \in \{1,2,3,4\}$, where we define the complexity level to be the number of depth with the corresponding reasoning tree to derive the queries. 
Specifically, $\mathcal{R}_{i,L}$ contains all nodes and edges in $\mathcal{R}_i$ that are within $L$ steps from the root $o_i$. 
For each level $L$, we generate implicit queries using the LLM:
\begin{equation}
(Q_{i,L}, C_{i,L}) = \text{LLM}(\mathcal{J}_{\text{sampled}}, o_i, \mathcal{T}_i, \mathcal{R}_{i,L}, L) \quad (i = 1, \cdots, N),
\end{equation}
where the LLM generates both the query $Q_{i,L}$ and identifies the corresponding reasoning categories $C_{i,L} \subseteq \{\text{semantic}, \text{spatial}, \text{temporal}\}$ based on the reasoning dimensions involved. 
Finally, for reasoning segmentation and grounding, we extract the corresponding ground truth $\mathcal{Y}_{i,L}$ from $\mathcal{J}$ by executing the reasoning defined in $\mathcal{R}_{i,L}$; and use LLM to generate the ground truth for reasoning summary and VQA.

\begin{figure*}[t!]
\centering
\includegraphics[width=\linewidth]{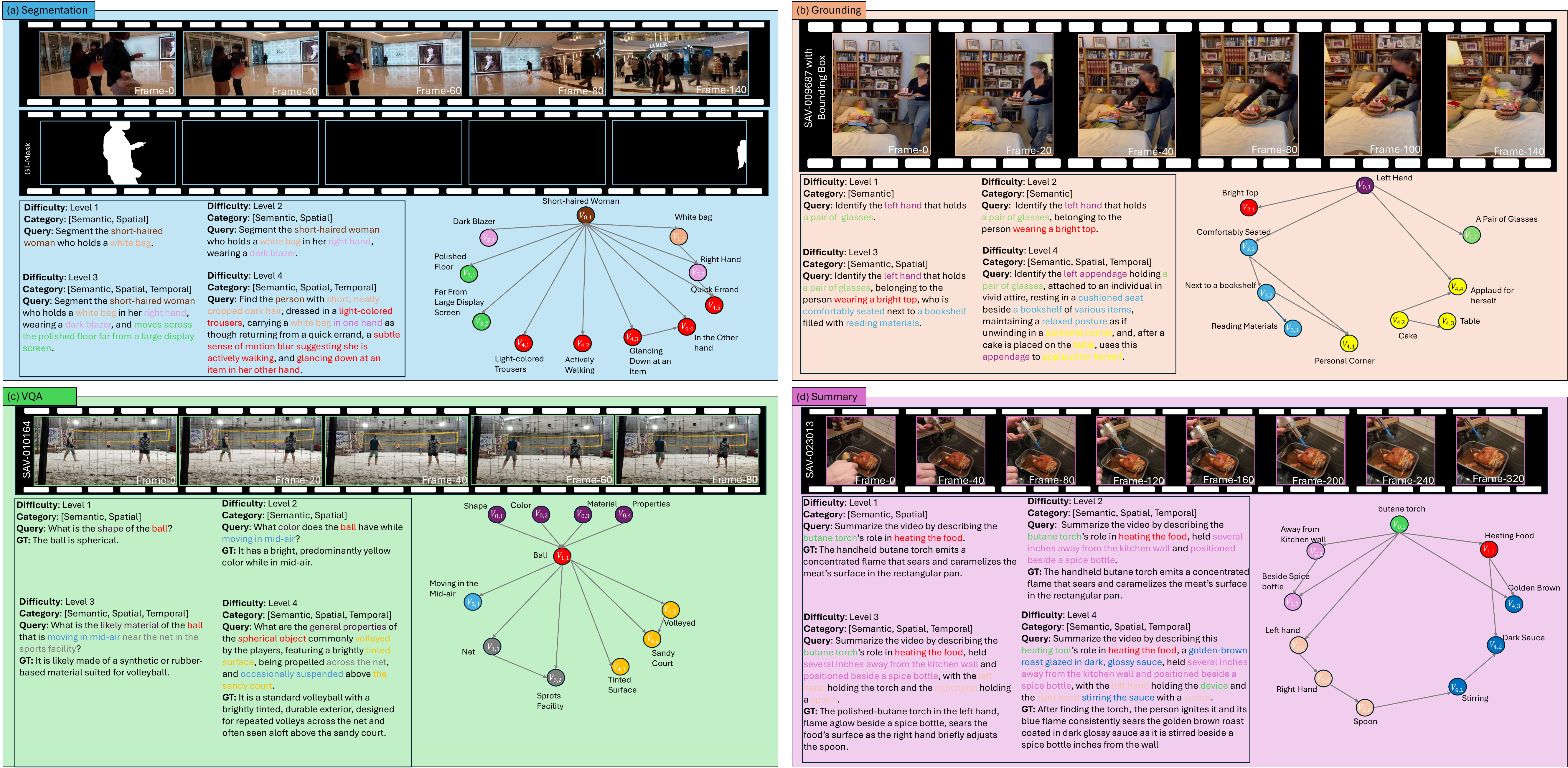}
\caption{
Examples of RVTBench across all four task types and difficulty levels. 
(a) Reasoning segmentation.
(b) Reasoning grounding. 
(c) Reasoning VQA. 
(d) Reasoning summary. 
For each example, we also demonstrate the reasoning tree accordingly (lower right of each panel), where nodes represent visual elements and edges indicate relationships between them. 
The complexity increases from level 1 to level 4 through the incorporation of additional semantic attributes, spatial relationships, and temporal dynamics.
}\label{fig:example}
\end{figure*}

\subsection{Dataset Statistics}

The RVTBench consists of a collection of RVTs in four types of tasks, three categories of reasoning, and four difficulty levels. 
In total, it contains 1,215,642 tokens distributed in 3,896 queries, with an increasing proportion of higher difficulty levels, where Level 4 accounting for 42.53\% of all tokens, followed by Level 3 (30.45\%), Level 2 (18.51\%), and Level 1 (8.51\%).
This progressive distribution ensures that the models are evaluated for their ability to handle increasingly complex reasoning chains.
In terms of task distribution, grounding tasks represent the largest portion at 43.52\% of the total tokens, followed by summary tasks (23.49\%), segmentation tasks (18.27\%), and VQA tasks (14.72\%), which reflects the varying complexity and verbosity required for different output modalities as grounding tasks typically requiring more detailed spatial specifications compared to other tasks.
Across reasoning categories, semantic reasoning dominates with 43.99\% of tokens, followed by spatial reasoning (36.26\%) and temporal reasoning (19.75\%).
The smaller proportion of temporal reasoning queries reflects the inherently higher complexity of temporal relationships, which are concentrated in the higher difficulty levels.
We show representative examples across all combinations of task types, reasoning categories, and difficulty levels, as illustrated in Fig.~\ref{fig:stats}.
Level 1 queries typically involve simple attribute identification (e.g., ``\textit{Segment the bear with a thick, shaggy coat}''), while level 2 queries introduce basic relationships (e.g., ``\textit{Segment the white cargo vehicle on the left edge of the roadway}'').
Level 3 and 4 queries progressively incorporate more complex reasoning chains, such as ``\textit{Identify the left arm of the blonde-haired woman standing behind the glasses-wearing woman at the entrance of the game store}'', which combines multiple spatial relationships and semantic attributes.
Temporal reasoning queries, which are found in higher difficulty levels, include examples like ``\textit{Segment the man who will rotate the bicycle}'' and ``\textit{Summarize the video by describing the boy riding a bicycle, who later passes a pile of parked bikes}'', requiring understanding of the sequence and events across video frames.
Fig. \ref{fig:example} illustrates representative examples in all types of RVT and difficulty levels in RVTBench.

\begin{table*}[!htbp]
\caption{Performance comparison on reasoning segmentation queries from RVTBench, where our proposed RVTagent demonstrates substantial improvements over existing approaches. ``ft'' stands for fine-tuned model.}
\label{table:exp2}
\centering
\resizebox{\linewidth}{!}{%
  \begin{tabular}{
  l|
  cccc @{\hspace{1em}}
  cccc @{\hspace{1em}}
  cccc @{\hspace{1em}}
  |
  cccc @{\hspace{1em}}
  cccc @{\hspace{1em}}
  cccc @{\hspace{1em}}
  |
  cccc @{\hspace{1em}}
  cccc @{\hspace{1em}}
  cccc
  }
\toprule
& \multicolumn{12}{c|}{$\bm{\mathcal{J}}$ (\%) ($\uparrow$)}
& \multicolumn{12}{c|}{$\bm{\mathcal{F}}$ (\%) ($\uparrow$)}
& \multicolumn{12}{c}{$\bm{\mathcal{J\&F}}$ (\%) ($\uparrow$)} \\
\cmidrule(lr){2-13} \cmidrule(lr){14-25} \cmidrule(lr){26-37}
\textbf{Model}
& \multicolumn{4}{c}{Semantic}
& \multicolumn{4}{c}{Spatial}
& \multicolumn{4}{c|}{Temporal}
& \multicolumn{4}{c}{Semantic}
& \multicolumn{4}{c}{Spatial}
& \multicolumn{4}{c|}{Temporal}
& \multicolumn{4}{c}{Semantic}
& \multicolumn{4}{c}{Spatial}
& \multicolumn{4}{c}{Temporal} \\
\cmidrule(lr){2-5} \cmidrule(lr){6-9} \cmidrule(lr){10-13}
\cmidrule(lr){14-17} \cmidrule(lr){18-21} \cmidrule(lr){22-25}
\cmidrule(lr){26-29} \cmidrule(lr){30-33} \cmidrule(lr){34-37}
& L1 & L2 & L3 & L4
& L1 & L2 & L3 & L4
& L1 & L2 & L3 & L4
& L1 & L2 & L3 & L4
& L1 & L2 & L3 & L4
& L1 & L2 & L3 & L4
& L1 & L2 & L3 & L4
& L1 & L2 & L3 & L4
& L1 & L2 & L3 & L4 \\
\midrule
\textbf{LISA-7B \cite{lisa}}   & 10.77 & 10.86 & 11.06 & 10.43 & 10.24 & 10.09 & 10.60 & 10.26 & 21.59 & 13.97 & 13.63 & 11.29 
& 7.21 & 7.21 & 7.42 & 6.55 & 6.69 & 6.57 & 7.07 & 6.45 & 16.42 & 9.74 & 9.23 & 7.01 
& 8.99 & 9.04 & 9.24 & 8.49 & 8.47 & 8.33 & 8.83 & 8.36 & 19.06 & 11.86 & 11.43 & 9.15
\\
\textbf{LISA-13B \cite{lisa}}  & 11.49 & 11.79 & 11.82 & 10.89 & 11.49 & 10.69 & 11.35 & 10.61 & 22.80 & 16.08 & 14.14 & 11.81 
& 7.88 & 8.18 & 8.13 & 7.30 & 7.71 & 7.25 & 7.80 & 7.13 & 16.53 & 11.42 & 9.65 & 7.58 
& 9.69 & 9.98 & 9.97 & 9.09 & 9.60 & 8.97 & 9.57 & 8.87 & 19.66 & 13.75 & 11.89 & 9.70 \\
\textbf{LISA++ \cite{lisa++}}    & 10.34 & 10.49 & 10.79 & 9.87 & 9.93 & 9.71 & 10.44 & 9.54 & 24.79 & 15.07 & 13.07 & 11.10 
& 6.72 & 6.78 & 7.08 & 6.23 & 5.79 & 6.18 & 6.95 & 6.03 & 18.14 & 10.53 & 8.67 & 6.90 
& 8.53 & 8.63 & 8.94 & 8.05 & 7.86 & 7.94 & 8.69 & 7.79 & 21.46 & 12.80 & 10.87 & 9.00 \\
\textbf{VISA \cite{visa}}   & 3.94 & 3.55 & 3.34 & 2.97 & 4.54 & 3.89 & 3.77 & 3.13 & 7.41 & 2.93 & 2.95 & 2.57
& 8.89 & 9.80 & 9.56 & 9.02 & 10.13 & 9.79 & 9.06 & 8.74 & 13.08 & 10.34 & 9.13 & 9.54
& 6.42 & 6.53 & 6.45 & 6.00 & 7.34 & 6.84 & 6.42 & 5.94 & 10.25 & 6.67 & 6.04 & 6.06 \\
\textbf{SegLLM \cite{segllm}}      & 9.52 & 9.48 & 9.62 & 9.37 & 9.57 & 9.16 & 9.44 & 9.14 & 19.95 & 12.90 & 11.72 & 10.64
& 12.99 & 13.17 & 13.22 & 12.92 & 12.93 & 12.87 & 13.03 & 12.65 & 24.40 & 16.82 & 15.69 & 14.49
& 11.25 & 11.33 & 11.42 & 11.14 & 11.25 & 11.01 & 11.24 & 10.90 & 22.18 & 14.86 & 13.71 & 12.57 \\
\textbf{SAM4MLLM \cite{sam4mllm}}      & 10.37 & 11.38 & 11.54 & 11.88 & 11.63 & 11.05 & 11.14 & 11.53 & 15.89 & 13.23 & 13.34 & 13.05 
& 7.33 & 8.20 & 8.49 & 8.79 & 8.44 & 7.91 & 8.12 & 8.51 & 12.30 & 9.40 & 9.49 & 9.35 
& 8.85 & 9.79 & 10.02 & 10.34 & 10.03 & 9.48 & 9.63 & 10.02 & 14.10 & 11.31 & 11.42 & 11.20 \\
\textbf{LaSagnA \cite{lasagna}}     & 7.77 & 7.88 & 7.97 & 7.02 & 9.25 & 7.69 & 7.86 & 6.89 & 14.78 & 10.49 & 9.73 & 7.48
& 10.8 & 10.95 & 11.21 & 10.23 & 12.26 & 10.71 & 11.01 & 10.03 &20.75 & 14.26 & 13.38 & 11.22
& 9.29 & 9.42 & 9.59 & 8.63 & 10.75 & 9.20 & 9.43 & 8.46 & 17.76 & 12.37 & 11.56 & 9.35 \\
\midrule
\textbf{LISA-7B (ft) \cite{lisa}}   & 45.77 & 44.86 & 42.06 & 40.43 & 43.24 & 42.09 & 40.60 & 38.26 & 41.59 & 38.97 & 36.63 & 34.29  & 42.21 & 41.21 & 39.42 & 36.55 & 40.69 & 39.57 & 38.07 & 35.45 & 39.42 & 35.74 & 33.23 & 31.01 & 43.99 & 43.04 & 40.74 & 38.49 & 41.97 & 40.83 & 39.34 & 36.86 & 40.51 & 37.36 & 34.93 & 32.65 \\
\textbf{LISA-13B (ft) \cite{lisa}}  & 48.49 & 47.79 & 45.82 & 43.89 & 46.49 & 45.69 & 44.35 & 42.61 & 44.80 & 42.08 & 39.14 & 36.81 & 45.88 & 45.18 & 43.13 & 41.30 & 44.71 & 43.25 & 41.80 & 40.13 & 42.53 & 39.42 & 36.65 & 34.58 & 47.19 & 46.49 & 44.48 & 42.60 & 45.60 & 44.47 & 43.08 & 41.37 & 43.67 & 40.75 & 37.90 & 35.70 \\
\textbf{VISA (ft) \cite{lisa}} & 51.20 & 49.45 & 47.70 & 45.15 & 48.75 & 47.20 & 45.55 & 43.10 & 44.40 & 41.85 & 39.55 & 37.20 & 49.10 & 47.70 & 46.05 & 43.55 & 47.25 & 45.80 & 44.15 & 41.70 & 42.60 & 40.05 & 37.75 & 35.40 & 50.15 & 48.58 & 46.88 & 44.35 & 48.00 & 46.50 & 44.85 & 42.40 & 43.50 & 40.95 & 38.65 & 36.30 \\
\midrule
\textbf{RVTagent} & 74.80 & 72.65 & 70.30 & 67.95 & 72.40 & 70.15 & 67.85 & 65.40 & 68.75 & 66.20 & 63.90 & 61.55 
& 72.60 & 70.45 & 68.10 & 65.75 & 70.20 & 67.95 & 65.65 & 63.20 & 66.55 & 64.00 & 61.70 & 59.35 
& 73.70 & 71.55 & 69.20 & 66.85 & 71.30 & 69.05 & 66.75 & 64.30 & 67.65 & 65.10 & 62.80 & 60.45 \\
\bottomrule
\end{tabular}%
}
\end{table*}

\subsection{Proposed Baseline}
We propose RVTagent as a baseline method for RVTs that enables zero-shot generalization without requiring model fine-tuning. 
Formally, given an implicit text query $Q$, RVTagent first analyzes the query to determine the task type $\mathcal{T} \in \{\text{segmentation}, \text{grounding}, \text{summary}, \text{VQA}\}$ and constructs a reasoning strategy. 
This task identification process uses an LLM that builds a reasoning graph $\mathcal{R} = (V, E)$ in a zero-shot manner, where nodes $V$ represent atomic reasoning operations that will later be used to determine the model for the construction of the DT representation, and edges $E$ encode dependencies between them \cite{jit}. 
The planning process is formalized as $(\mathcal{T}, \mathcal{R}) = \text{LLM}(Q)$, where the LLM decomposes complex queries into a sequence of simpler operations tailored to the specific reasoning requirements.
Based on the identified task type and reasoning graph, RVTagent then constructs a task-specific DT representation. 
Formally, for a video sequence $\mathcal{X}$, it builds a representation $\mathcal{J} = \{\mathcal{J}^{(1)}, \mathcal{J}^{(2)}, \ldots, \mathcal{J}^{(T)}\}$ by selecting appropriate pre-trained models from HuggingFace based on the planning results, which ensures that only the most relevant visual features are extracted for the specific task requirements. 
%
%
This structured representation is organized as a JSON graph where nodes represent objects and edges encode relationships, preserving both semantic attributes and spatial-temporal dynamics important for reasoning.
In the final stage, RVTagent executes the reasoning graph on the DT representation to produce task-appropriate output. 
For each node of reasoning $v_i \in V$, the corresponding operation is applied by $y_i = f_i(x_i; \theta_i)$, where $x_i$ represents the input features of the predecessor nodes, $\theta_i$ denotes the operation parameters, and $f_i$ is the reasoning function implemented by the LLM. 
Finally, the output format is determined by the type of identified task. 

\section{Experiments}

\paragraph{Implementation Details}
All experiments were carried out with Python 3.10.16 and PyTorch 2.1.2 on 8 NVIDIA RTX 4090 GPUs with 24GB memory. 
For the construction of the DT representation, we processed key frames with $t_s$ based on video length and object numbers to balance computational efficiency with temporal coherence. 
For generating implicit queries and reasoning trees, we employed OpenAI's 4o with temperature set to 0.7 and top-p to 0.95 to ensure appropriate diversity in query generation while maintaining coherence. 
%
%
We evaluate reasoning segmentation using the Jaccard index ($\mathcal{J}$) \cite{regionsim} and F-measure ($\mathcal{F}$) \cite{contouracc}, with their mean as $\mathcal{J}\&\mathcal{F}$\cite{j&f}. 
For grounding, we compute both the cumulative Intersection over Union (cIoU) \cite{lisa}  and the average per-image IoU (gIoU) \cite{lisa,giou}, along with average precision at an IoU threshold of 0.5 (AP@50). 
For summary and VQA, we assess token-level overlap using BLEU-4 \cite{bleu} and ROUGE-L \cite{rouge}, semantic similarity with BERTScore \cite{bertscore}, and consensus-based evaluation via CIDEr \cite{cider}.

\begin{table*}[!htbp]
\caption{Performance comparison on reasoning grounding task with RVTBench.}\label{table:exp3}
\centering
\resizebox{\linewidth}{!}{%
\begin{tabular}{
l|
cccc @{\hspace{1em}}
cccc @{\hspace{1em}}
cccc @{\hspace{1em}}
|
cccc @{\hspace{1em}}
cccc @{\hspace{1em}}
cccc @{\hspace{1em}}
|
cccc @{\hspace{1em}}
cccc @{\hspace{1em}}
cccc
}
\toprule
& \multicolumn{12}{c|}{cIoU (\%) ($\uparrow$)}
& \multicolumn{12}{c|}{gIoU (\%) ($\uparrow$)}
& \multicolumn{12}{c}{AP@50 (\%) ($\uparrow$)} \\
\cmidrule(lr){2-13} \cmidrule(lr){14-25} \cmidrule(lr){26-37}
\textbf{Model}
& \multicolumn{4}{c}{Semantic}
& \multicolumn{4}{c}{Spatial}
& \multicolumn{4}{c|}{Temporal}
& \multicolumn{4}{c}{Semantic}
& \multicolumn{4}{c}{Spatial}
& \multicolumn{4}{c|}{Temporal}
& \multicolumn{4}{c}{Semantic}
& \multicolumn{4}{c}{Spatial}
& \multicolumn{4}{c}{Temporal} \\
\cmidrule(lr){2-5} \cmidrule(lr){6-9} \cmidrule(lr){10-13}
\cmidrule(lr){14-17} \cmidrule(lr){18-21} \cmidrule(lr){22-25}
\cmidrule(lr){26-29} \cmidrule(lr){30-33} \cmidrule(lr){34-37}
& L1 & L2 & L3 & L4
& L1 & L2 & L3 & L4
& L1 & L2 & L3 & L4
& L1 & L2 & L3 & L4
& L1 & L2 & L3 & L4
& L1 & L2 & L3 & L4
& L1 & L2 & L3 & L4
& L1 & L2 & L3 & L4
& L1 & L2 & L3 & L4 \\
\midrule
\textbf{LISA-7B \cite{lisa}}   & 8.58 & 8.71 & 9.09 & 8.89 & 8.68 & 8.52 & 8.73 & 8.85 & 10.02 & 10.29 & 9.64 & 8.97 
& 9.28 & 9.37 & 9.06 & 9.11 & 9.62 & 9.46 & 9.39 & 9.07 & 11.67 & 10.51 & 10.30 & 9.02 
& 6.03 & 6.20 & 6.41 & 5.70 & 6.42 & 6.06 & 6.04 & 5.63 & 8.35 & 7.67 & 7.12 & 5.78 \\
\textbf{LISA-13B \cite{lisa}}  & 9.22 & 9.24 & 9.66 & 8.97 & 9.55 & 9.19 & 9.32 & 8.86 & 11.04 & 11.40 & 10.21 & 9.01 
& 9.81 & 10.01 & 10.10 & 9.37 & 10.13 & 10.01 & 9.94 & 9.34 & 11.79 & 11.36 & 10.36 & 9.45 
& 6.67 & 6.96 & 6.93 & 5.99 & 7.23 & 7.05 & 6.89 & 5.91 & 9.60 & 8.61 & 7.02 & 6.16 \\
\textbf{LISA++ \cite{lisa++}}  & 8.68 & 9.18 & 9.56 & 9.62 & 8.77 & 8.84 & 9.07 & 9.74 & 13.69 & 11.09 & 9.76 & 10.40 
& 9.15 & 9.14 & 9.53 & 8.70 & 9.21 & 9.18 & 9.44 & 8.72 & 12.40 & 10.63 & 10.02 & 9.21 
& 6.13 & 6.53 & 6.88 & 5.92 & 6.21 & 6.62 & 6.79 & 5.96 & 9.60 & 8.07 & 7.69 & 6.39 \\
\textbf{VISA \cite{visa}}   & 13.30 & 14.41 & 14.79 & 14.47 & 12.37 & 14.48 & 15.98 & 14.41 & 4.67 & 16.20 & 14.30 & 15.48
& 14.97 & 16.05 & 15.57 & 16.25 & 15.28 & 15.48 & 16.05 & 16.40 & 13.25 & 17.08 & 16.09 & 15.97
& 7.50 & 9.10 & 8.47 & 8.91 & 9.47 & 8.21 & 8.84 & 9.05 & 7.28 & 7.65 & 9.12 & 8.58 \\
\textbf{SegLLM \cite{segllm}}      & 8.94 & 9.77 & 10.01 & 10.60 & 10.07 & 9.52 & 9.64 & 10.69 & 13.51 & 11.81 & 10.85 & 11.90
& 11.07 & 11.39 & 11.47 & 11.44 & 11.70 & 11.53 & 11.51 & 11.40 & 15.19 & 12.67 & 12.47 & 11.84
& 9.60 & 9.81 & 9.98 & 9.68 & 10.30 & 9.88 & 10.05 & 9.60 & 14.78 & 11.45 & 11.14 & 10.14 \\
\textbf{SAM4MLLM \cite{sam4mllm}}      & 8.67 & 9.39 & 9.43 & 9.77 & 9.04 & 9.35 & 8.90 & 9.74 & 6.75 & 7.88 & 9.53 & 10.68 
& 10.85 & 11.26 & 11.52 & 11.71 & 10.92 & 11.29 & 11.33 & 11.57 & 12.52 & 12.18 & 11.98 & 12.47 
& 8.27 & 8.90 & 9.19 & 9.26 & 8.38 & 8.78 & 8.99 & 9.09 & 11.16 & 10.17 & 9.74 & 10.18 \\

\textbf{LaSagnA \cite{lasagna}}     & 10.18 & 9.92 & 10.01 & 9.25 & 10.06 & 9.92 & 9.75 & 9.22 & 10.52 & 10.47 & 10.39 & 9.71
& 10.55 & 10.26 & 10.14 & 9.38 & 10.65 & 10.36 & 10.21 & 9.43 & 12.45 & 11.55 & 11.06 & 9.68
& 8.03 & 7.85 & 7.83 & 6.44 & 8.16 & 8.07 & 7.79 & 6.49 & 9.50 & 9.30 & 8.46 & 6.72 \\
\midrule
\textbf{LISA-7B (ft) \cite{lisa}}   & 42.85 & 41.20 & 39.45 & 37.60 & 40.30 & 38.85 & 37.20 & 35.45 & 36.65 & 35.10 & 33.25 & 31.70 
& 40.95 & 39.30 & 37.55 & 35.70 & 38.40 & 36.95 & 35.30 & 33.55 & 34.75 & 33.20 & 31.35 & 29.80 
& 38.45 & 36.80 & 35.05 & 33.20 & 35.90 & 34.45 & 32.80 & 31.05 & 32.25 & 30.70 & 28.85 & 27.30 \\
\textbf{LISA-13B (ft) \cite{lisa}}  & 45.95 & 44.30 & 42.55 & 40.70 & 43.40 & 41.95 & 40.30 & 38.55 & 39.75 & 38.20 & 36.35 & 34.80 
& 44.05 & 42.40 & 40.65 & 38.80 & 41.50 & 40.05 & 38.40 & 36.65 & 37.85 & 36.30 & 34.45 & 32.90 
& 41.55 & 39.90 & 38.15 & 36.30 & 39.00 & 37.55 & 35.90 & 34.15 & 35.35 & 33.80 & 31.95 & 30.40 \\
\textbf{VISA (ft) \cite{lisa}} & 49.05 & 47.40 & 45.65 & 43.80 & 46.50 & 45.05 & 43.40 & 41.65 & 42.85 & 41.30 & 39.45 & 37.90 
& 47.15 & 45.50 & 43.75 & 41.90 & 44.60 & 43.15 & 41.50 & 39.75 & 40.95 & 39.40 & 37.55 & 36.00 
& 44.65 & 43.00 & 41.25 & 39.40 & 42.10 & 40.65 & 39.00 & 37.25 & 38.45 & 36.90 & 35.05 & 33.50 \\
\midrule
\textbf{RVTagent} & 69.85 & 67.40 & 65.25 & 62.90 & 67.30 & 65.05 & 62.80 & 60.55 & 63.75 & 61.50 & 59.35 & 57.10 
& 67.95 & 65.50 & 63.35 & 61.00 & 65.40 & 63.15 & 60.90 & 58.65 & 61.85 & 59.60 & 57.45 & 55.20 
& 65.45 & 63.00 & 60.85 & 58.50 & 62.90 & 60.65 & 58.40 & 56.15 & 59.35 & 57.10 & 54.95 & 52.70 \\
\bottomrule
\end{tabular}%
}
\end{table*}

\begin{table*}[!htbp]
\caption{Performance comparison on reasoning VQA and summary tasks with RVTBench.}\label{table:exp4}
\centering
\newcommand{\fourPH}{\plh & \plh & \plh & \plh}
\resizebox{\linewidth}{!}{%
\begin{tabular}{
l|l|
*{3}{cccc@{\hspace{1em}}}|
*{3}{cccc@{\hspace{1em}}}|
*{3}{cccc@{\hspace{1em}}}|
*{3}{cccc}
}
\toprule
& & \multicolumn{12}{c|}{\textbf{BLEU-4} (\%) ($\uparrow$)}
& \multicolumn{12}{c|}{\textbf{ROUGE-L} (\%) ($\uparrow$)}
& \multicolumn{12}{c|}{\textbf{BertScore} (\%) ($\uparrow$)}
& \multicolumn{12}{c}{\textbf{CIDEr} (\%) ($\uparrow$)} \\
\cmidrule(lr){3-14}  \cmidrule(lr){15-26}
\cmidrule(lr){27-38} \cmidrule(lr){39-50}
\textbf{Task} & \textbf{Model}
& \multicolumn{4}{c}{\textbf{Semantic}}
& \multicolumn{4}{c}{\textbf{Spatial}}
& \multicolumn{4}{c|}{\textbf{Temporal}}
& \multicolumn{4}{c}{\textbf{Semantic}}
& \multicolumn{4}{c}{\textbf{Spatial}}
& \multicolumn{4}{c|}{\textbf{Temporal}}
& \multicolumn{4}{c}{\textbf{Semantic}}
& \multicolumn{4}{c}{\textbf{Spatial}}
& \multicolumn{4}{c|}{\textbf{Temporal}}
& \multicolumn{4}{c}{\textbf{Semantic}}
& \multicolumn{4}{c}{\textbf{Spatial}}
& \multicolumn{4}{c}{\textbf{Temporal}} \\
\cmidrule(lr){3-6}   \cmidrule(lr){7-10}  \cmidrule(lr){11-14}
\cmidrule(lr){15-18} \cmidrule(lr){19-22} \cmidrule(lr){23-26}
\cmidrule(lr){27-30} \cmidrule(lr){31-34} \cmidrule(lr){35-38}
\cmidrule(lr){39-42} \cmidrule(lr){43-46} \cmidrule(lr){47-50}
& & 
L1 & L2 & L3 & L4
& L1 & L2 & L3 & L4
& L1 & L2 & L3 & L4
& L1 & L2 & L3 & L4
& L1 & L2 & L3 & L4
& L1 & L2 & L3 & L4
& L1 & L2 & L3 & L4
& L1 & L2 & L3 & L4
& L1 & L2 & L3 & L4
& L1 & L2 & L3 & L4
& L1 & L2 & L3 & L4
& L1 & L2 & L3 & L4 \\
\midrule

VQA & GPT-o4-mini
& 2.25 & 1.15 & 0.80
& 0.42 & 1.63 & 1.13
& 0.90 & 0.45 & 0.00
& 1.21 & 0.58 & 0.39

& 29.99 & 22.07 & 20.23
& 19.02 & 29.38 & 21.89
& 20.41 & 19.25 & 26.82
& 21.30 & 19.61 & 18.74

& 74.25 & 71.57 & 68.98 
& 69.04 & 72.87 & 71.40
& 69.16 & 69.28 & 72.20
& 71.34 & 68.80 & 69.34

& 70.76 & 49.97 & 25.50 
& 9.74 & 46.92 & 43.33
& 26.44 & 10.36 & 56.07
& 31.16 & 20.39 & 10.87
\\

VQA & Gemini-2.0-flash-lite
& 1.13 & 0.76 & 0.37
& 0.20 & 1.54 & 0.63
& 0.45 & 0.20 & 0.00
& 0.33 & 0.30 & 0.13

& 21.27 &17.24 & 14.61
& 13.99 & 23.04 & 17.38
& 14.88 & 14.35 & 22.21
& 15.96 & 14.63 & 13.50

& 65.28 & 63.20 & 57.90
& 54.17 & 63.96 & 62.67
& 57.97 & 54.38 & 65.03
& 60.35 & 57.29 & 53.77

& 34.05 & 24.72 & 5.55
& 1.07 & 33.84 & 21.18
& 6.34 & 1.34 & 46.97
& 14.98 & 5.11 & 0.88\\

VQA & Claude3-Haiku
& 0.94 & 0.66 & 0.35
& 0.33 & 0.83 & 0.60
& 0.36 & 0.35 & 0.00
& 0.27 & 0.33 & 0.21

& 18.48 & 16.44 & 13.72
& 13.80 & 19.36 & 16.31
& 13.97 & 14.03 & 12.63
& 14.42 & 13.45 & 12.92

& 65.41 & 65.09 & 62.44
& 62.31 & 65.42 & 64.86
& 62.71 & 62.53 & 62.34
& 65.31 & 62.31 & 62.04

& 9.62 & 8.84 & 0.83
& 0.13 & 0.80 & 5.44
& 1.03 & 0.17 & 0.00
& 1.29 & 0.46 & 0.03\\

VQA & Qwen2.5-omni \cite{Qwen2.5-Omni}
& 0.89 & 0.48 & 0.40
& 0.41 & 1.28 & 0.59
& 0.44 & 0.42 & 0.00
& 0.53 & 0.45 & 0.35

& 17.72 & 13.44 & 12.71
& 13.64 & 20.73 & 13.48
& 12.89 & 13.88 & 10.48
& 12.99 & 12.42 & 13.08

& 61.74 & 59.57 & 57.60
& 58.39 & 62.43 & 59.48
& 57.86 & 58.52 & 55.65
& 60.16 & 57.64 & 58.40

& 6.09 & 3.35 & 1.25
& 0.35 & 8.09 & 3.84
& 1.21 & 0.43 & 0.00
& 1.90 & 0.93 & 0.22 \\

VQA & Janus-Pro-7B \cite{janus-pro}
& 2.92 & 1.72 & 1.30
& 1.18 & 3.52 & 1.87
& 1.47 & 1.30 & 0.00
& 1.15 & 1.23 & 1.06

& 28.40 & 21.58 & 19.23
& 19.36 & 30.04 & 21.59
& 19.70 & 19.92 & 20.60
& 19.81 & 19.14 & 18.92

& 72.32 & 69.99 & 67.12
& 66.50 & 71.29 & 69.93
& 67.48 & 66.72 & 70.55
& 69.04 & 66.71 & 66.29

& 74.46 & 54.38 & 22.17
& 10.25 & 62.72 & 53.62
& 23.69 & 11.02 & 22.47
& 35.36 & 17.67 & 9.28\\

VQA & LISA-7B \cite{lisa}
& 2.66 & 1.26 & 0.70 
& 0.54 & 2.32 & 1.52 
& 0.82 & 0.58 & 0.00 
& 0.58 & 0.77 & 0.43

& 24.97 & 19.15 & 14.24 
& 14.28 & 24.26 & 19.26 
& 14.49 & 14.56 & 15.05 
& 15.92 & 14.59 & 13.97 

& 70.01 & 67.22 & 61.69 
& 62.89 & 68.08 & 67.15 
& 61.86 & 63.01 & 66.79 
& 66.14 & 62.58 & 62.83

& 74.25 & 46.61 & 11.56 
& 0.69 & 51.09 & 45.80 
& 12.98 & 0.68 & 19.73 
& 28.93 & 10.28 & 0.67\\

VQA & LISA-13B \cite{lisa}
& 2.72 & 1.64 & 0.74 
& 0.39 & 2.82 & 1.90 
& 0.88 & 0.39 & 0.00 
& 0.80 & 0.69 & 0.28

& 25.12 & 19.79 & 15.01 
& 13.52 & 23.59 & 19.75 
& 15.46 & 13.67 & 18.63 
& 17.71 & 14.91 & 13.08

& 70.80 & 68.60 & 63.34 
& 62.21 & 68.09 & 68.36 
& 63.73 & 62.21 & 68.96 
& 68.20 & 62.88 & 62.07

& 73.90 & 53.43 & 14.83 
& 0.41 & 49.55 & 51.78 
& 15.97 & 0.47 & 23.24 
& 33.67 & 9.08 & 0.45 \\

VQA & LISA\texttt{++} \cite{lisa++}
& 2.50 & 1.24 & 0.75 
& 0.46 & 3.03 & 1.19 
& 0.91 & 0.49 & 0.00 
& 0.59 & 0.60 & 0.33

& 25.88 & 19.20 & 15.63 
& 14.47 & 28.39 & 19.01 
& 16.20 & 14.67 & 16.36 
& 16.43 & 14.70 & 13.87

& 70.55 & 67.67 & 63.35 
& 62.00 & 70.05 & 67.69 
& 63.82 & 62.05 & 67.68 
& 66.35 & 62.59 & 61.93

& 72.89 & 45.88 & 15.75 
& 3.78 & 55.86 & 42.75 
& 18.14 & 4.12 & 28.55 
& 27.28 & 11.91 & 4.12 \\

%
VQA & RVTagent
& 7.23 & 6.54 & 5.87 
& 5.10 & 6.84 & 6.18 
& 5.56 & 5.23 & 4.82 
& 5.65 & 5.08 & 4.76

& 42.86 & 39.75 & 37.40 
& 35.64 & 41.25 & 38.70 
& 36.85 & 35.72 & 39.58 
& 37.46 & 35.28 & 34.53 

& 84.37 & 82.60 & 80.42 
& 79.35 & 83.18 & 81.54 
& 79.87 & 78.95 & 82.40 
& 80.82 & 79.45 & 78.65

& 92.48 & 85.63 & 77.41 
& 71.28 & 88.35 & 81.94 
& 75.78 & 70.45 & 83.26 
& 79.52 & 73.40 & 68.75\\

\midrule
Summary & GPT-o4-mini
& 0.47 & 0.61 & 0.74
& 0.43 & 0.60 & 0.76
& 0.76 & 0.42 & 0.25
& 0.88 & 0.71 & 0.40 

& 20.67 & 20.66 & 21.41
& 21.22 & 21.11 & 20.90
& 21.32 & 21.12 & 18.43
& 20.45 & 20.74 & 20.61

& 68.73 & 69.14 & 69.33 
& 69.63 & 68.96 & 69.21
& 69.24 & 69.58 & 67.18
& 68.59 & 68.36 & 68.88

& 7.96 & 6.61 & 6.45 
& 4.71 & 7.48 & 6.01
& 6.13 & 4.41 & 1.86
& 5.38 & 2.94 & 3.24\\

Summary & Gemini-2.0-flash-lite
& 0.62 & 0.69 & 0.79
& 0.42 & 0.59 & 0.71
& 0.79 & 0.42 & 0.32
& 0.70 & 0.76 & 0.40

& 18.72 & 18.61 & 19.90
& 17.53 & 18.36 & 18.51
& 19.85 & 17.45 & 17.77
& 17.75 & 19.34 & 17.41

& 61.16 & 61.45 & 62.52
& 60.26 & 61.28 & 61.28
& 62.32 & 60.22 & 60.34
& 60.90 & 61.78 & 59.71

& 3.55 & 2.73 & 1.73
& 0.29 & 2.86 & 1.85
& 1.55 & 0.30 & 0.61
& 1.72 & 0.68 & 0.30\\

Summary & Claude3-Haiku
& 0.38 & 0.35 & 0.32
& 0.21 & 0.38 & 0.38
& 0.35 & 0.21 & 0.24
& 0.38 & 0.30 & 0.20

& 14.75 & 13.79 & 13.99
& 14.27 & 14.70 & 14.00
& 14.05 & 14.25 & 13.78
& 13.39 & 13.51 & 14.26

& 63.21 & 61.53 & 61.22
& 61.56 & 63.42 & 61.65
& 61.21 & 61.52 & 63.31
& 61.34 & 60.59 & 61.20

& 0.01 & 0.02 & 0.00
& 0.00 & 0.01 & 0.02
& 0.00 & 0.00 & 0.00
& 0.00 & 0.00 & 0.00\\

Summary & Qwen2.5-omni \cite{Qwen2.5-Omni}
& 0.50 & 0.86 & 1.07
& 0.98 & 0.52 & 0.87
& 1.08 & 0.98 & 0.43
& 0.86 & 1.03 & 0.98

& 17.00 & 17.41 & 19.13
& 20.17 & 16.88 & 17.46
& 19.14 & 20.10 & 16.33
& 16.72 & 18.63 & 20.14

& 60.75 & 61.34 & 61.83
& 63.28 & 60.85 & 61.23
& 61.83 & 63.26 & 60.78
& 60.83 & 61.38 & 63.45

& 1.91 & 1.61 & 2.11
& 0.53 & 1.38 & 1.64
& 1.89 & 0.54 & 0.81
& 0.57 & 1.76 & 0.49\\

Summary & Janus-Pro-7B \cite{janus-pro}
& 1.37 & 1.34 & 1.62
& 1.30 & 1.37 & 1.44
& 1.65 & 1.30 & 0.48
& 1.06 & 1.42 & 1.32

& 22.54 & 21.30 & 23.13
& 22.32 & 23.00 & 21.69
& 23.24 & 22.21 & 20.61
& 19.63 & 22.33 & 22.30

& 68.84 & 67.70 & 68.48
& 67.51 & 69.49 & 67.78
& 68.57 & 67.45 & 67.93
& 66.67 & 67.47 & 67.09

& 23.20 & 16.92 & 15.99
& 6.42 & 26.46 & 17.05
& 16.11 & 6.36 & 11.23
& 12.05 & 12.85 & 5.69\\

Summary & LISA-7B \cite{lisa}
& 0.33 & 0.35 & 0.45 
& 0.29 & 0.52 & 0.40 
& 0.47 & 0.28 & 0.41 
& 0.49 & 0.53 & 0.33

& 11.94 & 11.01 & 11.19 
& 8.58 & 14.56 & 11.76 
& 11.65 & 8.59 & 14.82 
& 13.28 & 12.71 & 9.54

& 58.06 & 57.01 & 56.87 
& 53.71 & 61.25 & 57.75 
& 57.23 & 53.72 & 62.23 
& 59.43 & 58.12 & 54.40

& 4.12 & 2.17 & 0.87 
& 0.01 & 7.18 & 2.34 
& 0.78 & 0.01 & 4.22 
& 2.50 & 1.10 & 0.01\\

Summary & LISA-13B \cite{lisa}
& 0.30 & 0.34 & 0.36 
& 0.20 & 0.31 & 0.36 
& 0.36 & 0.20 & 0.35 
& 0.39 & 0.39 & 0.20

& 11.51 & 12.46 & 12.19 
& 8.49 & 12.97 & 13.04 
& 12.41 & 8.48 & 12.65 
& 13.52 & 12.74 & 8.61

& 57.87 & 58.51 & 57.53 
& 53.75 & 59.62 & 59.02 
& 57.71 & 53.73 & 60.70 
& 59.60 & 58.83 & 53.59
& 0.72 & 0.65 & 0.25 & 0.01 
& 0.91 & 0.69 & 0.26 & 0.01 
& 0.59 & 0.17 & 0.33 & 0.01 \\

Summary & LISA\texttt{++} \cite{lisa++}
& 0.92 & 0.80 & 0.80 
& 0.36 & 0.95 & 0.81 
& 0.84 & 0.36 & 0.58 
& 0.57 & 0.71 & 0.38

&19.49 & 18.99 & 19.84 
& 13.98 & 20.26 & 19.16 
& 19.86 & 13.97 & 17.11 
& 16.80 & 18.78 & 13.94

& 66.62 & 66.00 & 66.35 
& 60.47 & 67.33 & 65.94
& 66.31 & 60.46 & 64.84 
& 64.33 & 64.97 & 60.11

& 14.83 & 9.17 & 8.88 
& 0.23 & 16.53 & 9.20 
& 9.01 & 0.26 & 6.50 
& 6.46 & 5.91 & 0.14 \\

Summary & RVTagent
& 5.68 & 5.23 & 4.95 
& 4.56 & 5.42 & 5.08 
& 4.83 & 4.50 & 4.95 
& 5.12 & 4.78 & 4.42

& 38.76 & 36.94 & 35.87 
& 34.25 & 37.85 & 36.18 
& 35.40 & 34.10 & 36.73 
& 35.42 & 34.86 & 33.94

& 82.45 & 80.87 & 79.35 
& 78.10 & 81.78 & 80.24 
& 78.86 & 77.95 & 80.63 
& 79.14 & 78.23 & 77.48

& 85.76 & 79.45 & 72.83 
& 67.40 & 82.54 & 76.32 
& 71.45 & 66.84 & 77.38 
& 73.52 & 70.18 & 65.72\\
\bottomrule
\end{tabular}%
}
\end{table*}

\paragraph{Results}
Table \ref{table:exp2} shows that our proposed RVTagent outperforms existing approaches (both zero-shot and fine-tuned ones) on reasoning segmentation tasks. 
Similarly, as shown in Table \ref{table:exp3}, RVTagent exhibits superior performance on reasoning grounding task, with cIoU scores consistently above 60\% across all reasoning categories and difficulty levels, surpassing fine-tuned counterparts that generally achieve 40-50\%. 
For reasoning VQA and summary tasks (Table \ref{table:exp4}), RVTagent demonstrates better capabilities with BLEU-4 scores more than twice as high as the strongest baseline models. 
Across all tasks, we observe that while fine-tuning existing models (indicated by ``ft'') yields substantial improvements over their zero-shot counterparts, RVTagent consistently delivers superior results by effectively bridging visual perception with higher-level reasoning processes.
Finally, we observe a consistent pattern of performance degradation as task difficulty increases from L1 to L4. 
Moreover, semantic reasoning generally yields the highest performance scores across all models and task types, followed by spatial reasoning, whereas temporal reasoning emerges as the most challenging category, particularly at higher difficulty levels, where models must track complex object interactions over time.

\section{Conclusion}

We introduce reasoning visual tasks (RVTs) as a unified formulation that generalizes visual reasoning across multiple output formats including segmentation masks, bounding boxes, natural language descriptions, and question-answer pairs. 
Correspondingly, we propose a novel automated benchmark construction pipeline leveraging DT representations as structured intermediaries between visual perception and high-level reasoning, overcoming the limitations of token-based generation approaches based on VLM that inadequately capture complex spatial-temporal relationships. 
Based on this method, we presented RVTBench, a RVT benchmark containing 3,896 queries spanning four task types, three reasoning categories, and four difficulty levels derived from 200 video sequences, together with a baseline method RVTagent.
Although our benchmark covers semantic, spatial, and temporal reasoning, it focuses primarily on physical attributes and relationships rather than abstract concepts or causal reasoning, which therefore points out a promising future direction.
Additionally, exploring how models can leverage DT representations directly during inference, rather than just for benchmark construction, may lead to another promising direction.

\newpage
\bibliographystyle{plain}
\bibliography{ref.bib}


\end{document}